\documentclass[10pt,twocolumn,letterpaper]{article}

\usepackage{cvpr}              % To produce the CAMERA-READY version
\usepackage{multirow} %
\definecolor{cvprblue}{rgb}{0.21,0.49,0.74}
\usepackage[pagebackref,breaklinks,colorlinks,allcolors=cvprblue]{hyperref}

\def\paperID{14635} % *** Enter the Paper ID here
\def\confName{CVPR}
\def\confYear{2025}

\title{ROS-SAM: High-Quality Interactive Segmentation for Remote Sensing \\ Moving Object}

\author{Zhe Shan$^{1}$, Yang Liu$^{2}$, Lei Zhou$^{1,*\,}$, Cheng Yan$^{3}$, Heng Wang$^{1}$, Xia Xie$^{1,*\,}$\\
	$^1$ Hainan University \quad $^2$ Zhejiang University \quad $^3$ Tianjin University}
    
\begin{document}
\maketitle

\footnotetext[1]{Corresponding author: Lei Zhou (leizhou@hainanu.edu.cn), Xia Xie (shelicy@hainanu.edu.cn)}

\begin{abstract}
The availability of large-scale remote sensing video data underscores the importance of high-quality interactive segmentation. However, challenges such as small object sizes, ambiguous features, and limited generalization make it difficult for current methods to achieve this goal. In this work, we propose ROS-SAM, a method designed to achieve high-quality interactive segmentation while preserving generalization across diverse remote sensing data. The ROS-SAM is built upon three key innovations: 1) LoRA-based fine-tuning, which enables efficient domain adaptation while maintaining SAM’s generalization ability, 2) Enhancement of deep network layers to improve the discriminability of extracted features, thereby reducing misclassifications, and 3) Integration of global context with local boundary details in the mask decoder to generate high-quality segmentation masks. Additionally, we design the data pipeline to ensure the model learns to better handle objects at varying scales during training while focusing on high-quality predictions during inference. Experiments on remote sensing video datasets show that the redesigned data pipeline boosts the IoU by 6\%, while ROS-SAM increases the IoU by 13\%. Finally, when evaluated on existing remote sensing object tracking datasets, ROS-SAM demonstrates impressive zero-shot capabilities, generating masks that closely resemble manual annotations. These results confirm ROS-SAM as a powerful tool for fine-grained segmentation in remote sensing applications. Code is available at: https://github.com/ShanZard/ROS-SAM.
\end{abstract}    
\section{Introduction}
\label{sec:intro}
%Thanks to developments in aerospace technology, satellites can already achieve to gaze at specific areas and to image large areas continuously~\cite{10130311,li2023recent}. A series of video remote sensing satellites, such as SkySat, Jinlin-1, OVS, and Loujia3, have returned a large volume of remote sensing video data~\cite{CHEN2024212}. Compared to static remote sensing images, video data has greater potential for applications in transportation, security, emergency response, environmental monitoring, etc. Therefore, the study of video satellite vision tasks is one of the frontiers and hottest in remote sensing today.

%Advancements in aerospace technology have enabled satellites to precisely focus on specific areas and continuously capture large-scale imagery~\cite{10130311,li2023recent}. A range of video remote sensing satellites, including SkySat, Jinlin-1, OVS, and Loujia-3, have generated vast amounts of video data~\cite{CHEN2024212}.
%Unlike static remote sensing images, video data offers enhanced potential for applications in fields such as transportation, security, emergency response, and environmental monitoring. As a result, research into video satellite vision tasks has emerged as one of the most promising and dynamic frontiers in remote sensing today. 

\begin{figure}
   \centering
   \includegraphics[width=1.0\linewidth]{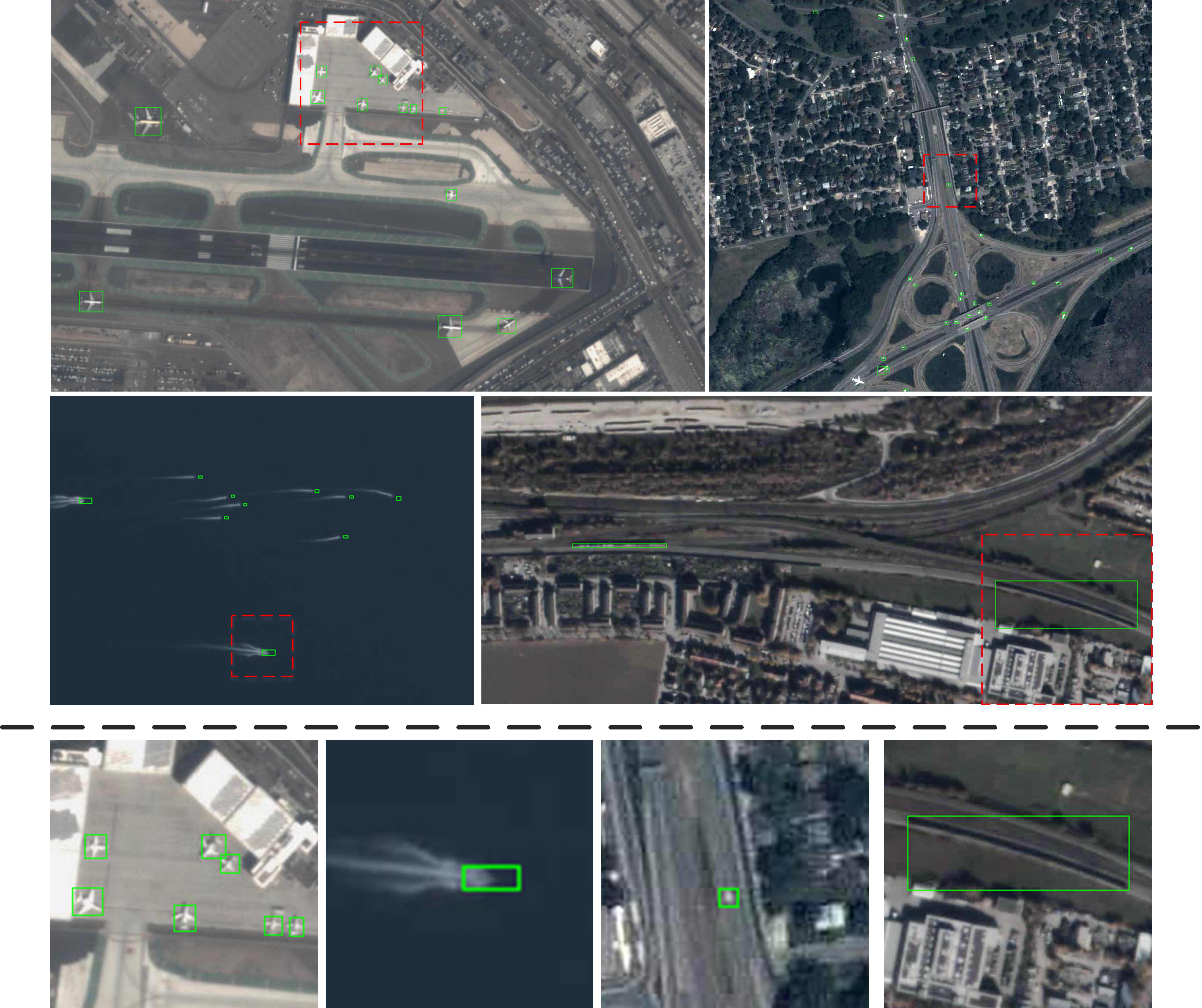}
   \caption{The recognition and annotation of these remote sensing video moving objects are particularly challenging due to factors such as small object sizes, lack of distinct features, sparse distribution, and other complexities.}
   \label{fig:1}
     \vspace{-15pt}  
\end{figure}

Research into video satellite vision tasks has emerged as one of the most promising and dynamic frontiers in remote sensing today~\cite{10130311,li2023recent,CHEN2024212}. However, for remote sensing video moving objects (RSVMO), the small size, ambiguous features, high object density, and the complexity and cost of frame-by-frame annotation make recognizing and segmenting these objects exceedingly difficult. As illustrated in Figure~\ref{fig:1}, airplanes, cars, ships, and trains are the most common moving objects in video satellite data, and both labeling these objects and developing algorithms to detect them are formidable challenges. At the same time, the lack of pixel-level labeling information directly leads to many algorithms that cannot be trained and inferred properly.

Remote sensing object tracking has been actively developed for years, with many labeled datasets already available~\cite{9625976,9533178,9715124,9672083,9875020,CHEN2024212}. It is natural to ask whether remote sensing video object segmentation can be achieved through interactive segmentation using existing object tracking datasets. Furthermore, the primary advantage of this approach lies in its ability to transform existing object tracking datasets into segmentation data, thereby advancing remote sensing video segmentation with minimal additional cost.

The Segment Anything Model (SAM)~\cite{Kirillov_2023_ICCV} is a foundational model designed for visual segmentation in images, trained on 1.1 billion diverse high-quality segmentation masks. This extensive training endows SAM with powerful prior knowledge and zero-shot capabilities. Moreover, the prompt-based interactive learning strategy of SAM is well-suited for converting bounding boxes into segmentation masks. However, our experiments show that SAM is unable to achieve this conversion effectively in the context of remote sensing data. This raises the question: what distinguishes SAM’s performance in general vision tasks from its limitations in remote sensing? We aim to address this question from three perspectives:

\textbf{(i)} Numerous empirical studies have highlighted a significant gap between general vision models and remote sensing tasks~\cite{9782149,Manas_2021_ICCV,10531642,10126079,Bastani_2023_ICCV}. A clear example is that objects in remote sensing data are typically unaffected by gravity, and their orientation can be ambiguous. When predicting the orientation of airplanes, SAM often struggles to determine the direction the plane is facing, often predicting a generic four-pointed star. Additionally, SAM's ability to distinguish features of ground objects is limited, it struggles to differentiate between ships and waves or airplanes and boarding bridges.

%\textbf{(ii)} The segmentation mask of SAM is usually not precise enough, and there are two key problems: rough mask boundaries and broken masks~\cite{NEURIPS2023_5f828e38}. The mask decoder of SAM employs a modification of the Transformer decoder, which takes the image embedding and prompt tokens for final mask prediction. Using this decoder to predict the exact position of the object from the image embedding that is missing a lot of texture edges is not easy. These reasons make it difficult for SAM to output high-quality masks similar to hand labeling.

\textbf{(ii)} SAM's prediction masks often suffer two main issues: rough mask boundaries and fragmented masks~\cite{NEURIPS2023_5f828e38}. The mask decoder of SAM employs a modified Transformer architecture, which takes image embeddings and prompt tokens to generate the mask prediction. However, predicting the precise location of an object from the image embedding, which often lacks fine texture details and edge information, proves challenging. These factors make it difficult for SAM to directly produce high-quality segmentation masks that closely resemble those created through manual annotation.

\textbf{(iii)} SAM requires input images to be resized to a fixed resolution of 1024×1024~\cite{kato2024generalized}, which poses challenges when working with remote sensing data. Remote sensing images typically have large spatial dimensions, but the objects of interest are often small. As a result, downsampling can cause these objects to vanish entirely. To address these issues, a more suitable pipeline for both training and inference must be designed, one that better accommodates the unique characteristics of remote sensing imagery.

In light of these analyses, we propose an effective approach, ROS-SAM, to fine-tune SAM for \textbf{R}emote sensing video moving \textbf{O}bject \textbf{S}egmentation. In addition, we design a novel training and inference pipeline aimed at achieving high-quality segmentation. First, we use Low-Rank Adaptation (LoRA)~\cite{hu2022lora} to adjust the parameters of the image encoder, minimizing the domain gap between the pre-training data and remote sensing data. We also unfreeze the last block of the image encoder to enhance its ability to extract more discriminative image embeddings. Second, we build on the HQ-SAM~\cite{NEURIPS2023_5f828e38} to modify the mask decoder, integrating both early and final features from the network to improve the restoration and refinement of predictions. Furthermore, unlike HQ-SAM, we do not freeze the mask decoder of the original SAM. Instead, we apply alternating optimization to update the parameters of both the encoder and decoder. This approach strengthens SAM's applicability to RSVMO while retaining its zero-shot capabilities. Finally, we reconstruct the training and inference pipeline by incorporating large-scale jittering (LSJ)~\cite{ghiasi2021simple} and random rotations during training to capture objects at various scales. For inference, we design a center-cropping strategy based on prompts to improve segmentation quality.

The main contributions can be summarized as follows: 
\begin{enumerate}
    \item We propose ROS-SAM for achieving high-quality prediction by incorporating remote sensing special knowledge and optimizing the mask decoder.
    \item We propose a novel data pipeline that introduces more multi-scale objects during training while inference focuses on high-quality inference for a single object. 
    \item Experimental results demonstrate that our data pipeline and ROS-SAM significantly improve performance on the SAT-MTB dataset~\cite{10130311}, increasing IoU by 6\% and 13\% over the original SAM. Furthermore, ROS-SAM exhibits strong zero-shot capabilities when evaluated on existing remote sensing object tracking datasets.
\end{enumerate}

\begin{figure*}
  \centering
  \includegraphics[width=0.85\linewidth]{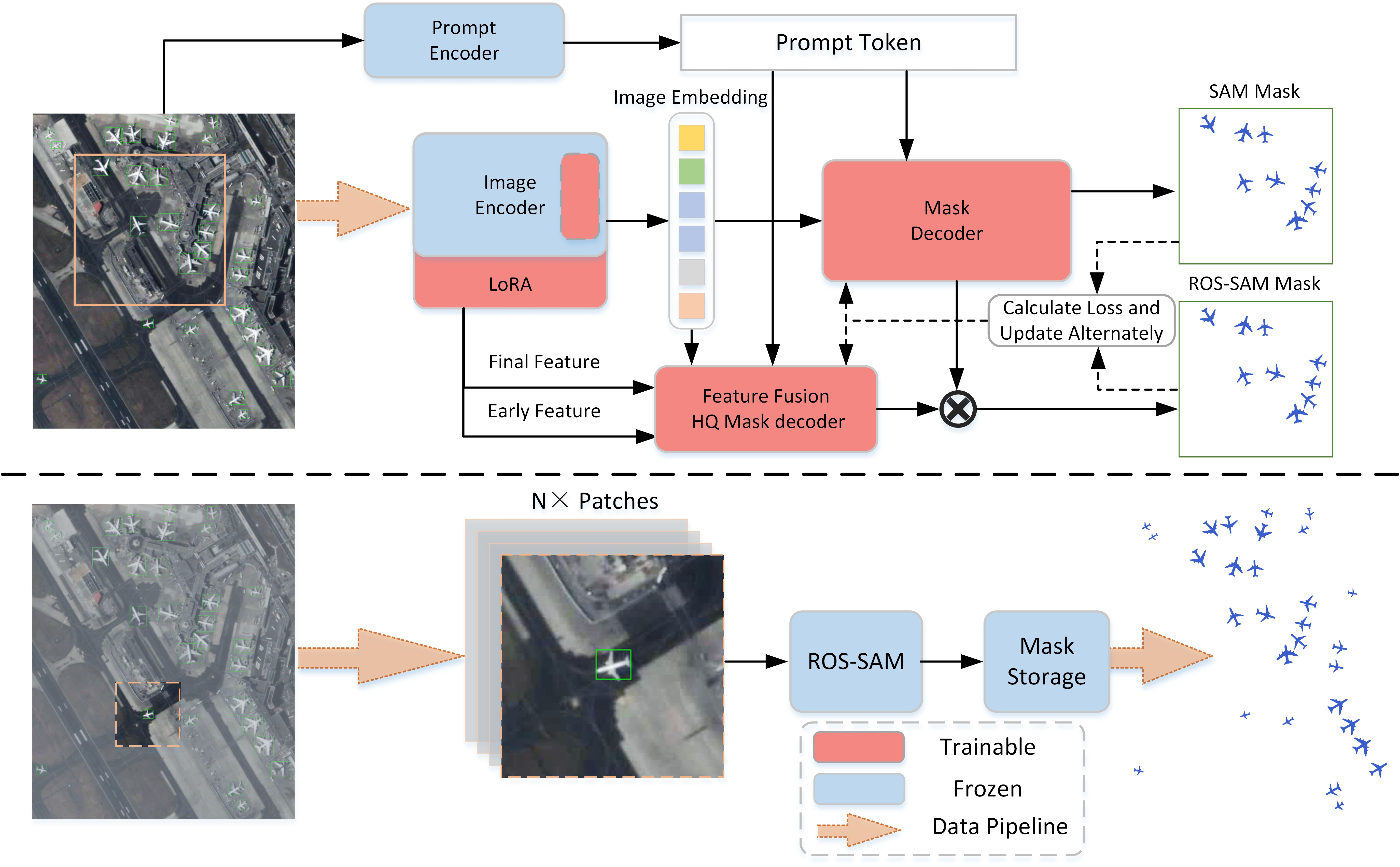}
  \caption{Overview of the proposed method. The upper part and the lower part outline the training and inference process, respectively.  
}
  \label{fig:2}
    \vspace{-15pt}  
\end{figure*}

\section{Related Work}

\textbf{Video remote sensing} is an emerging field within remote sensing, providing the foundation for real-time analysis of ground objects. The moving objects in remote sensing video differ significantly from those in images, exhibiting inherent characteristics such as low resolution, wide variations in scale, and high levels of noise. These factors pose significant challenges for vision algorithms in effectively analyzing the data. Early research in this field has primarily focused on three key areas: developing more robust feature extraction methods~\cite{shao2019can,shao2019tracking,xuan2021rotation,chen2022single}, establishing temporal relationships across consecutive frames~\cite{shao2019pasiam,song2022joint,yang2023siammdm}, and capturing spatiotemporal information to enable multi-frame associations~\cite{10632197,10638089}. These efforts have significantly enhanced the accuracy of object tracking and prompted the development of this field. Video object segmentation, however, is a more practical and complex task than object tracking. It requires the model to rely on prior knowledge, significant features, or manual prompts to identify regions of interest and perform recognition across continuous frames~\cite{9345705,9966836,10163641}. Despite the challenges, researchers have made initial attempts at segmenting objects in satellite video. One notable contribution is the SAT-MTB dataset~\cite{10130311}, a multitask benchmark that includes object tracking, detection, and segmentation tasks for satellite video. However, the dataset contains only 249 satellite videos and some of these lack annotations. With such a limited dataset, achieving high-quality remote sensing video segmentation is virtually impossible, underscoring the need for more comprehensive and well-annotated data in this area.

\textbf{Fine-tuning SAM} for application across various tasks has gained significant attention from the vision community. Numerous empirical studies have demonstrated that SAM's powerful prior knowledge can effectively guide a range of downstream tasks, including  medical image analysis~\cite{ma2024segment,MAZUROWSKI2023102918,HUANG2024103061,zhu2024medical}, 3D vision~\cite{shen2024micca,Yin_2024_CVPR,xu2024embodiedsam,chen2024sam}, video analysis~\cite{zhang2023uvosam,yang2023track,lu2023can,10418101}, remote sensing~\cite{OSCO2023103540,10315957,10636322,10680168}, and others~\cite{sun2024efficient,zhang2023attack,mo2023av,liu2024matcher}. The key challenge in fine-tuning SAM is to preserve its robust zero-shot capabilities while adapting it to new, task-specific objectives. Two notable methods for this adaptation are Adapter and LoRA. The Adapter~\cite{Chen_2023_ICCV,chen2024sam2} is a simple, flexible, and effective component that serves as an additional network layer, enabling the injection of task-specific guidance from a small amount of data. Its flexibility allows it to be combined with different parts of SAM, making it suitable for a variety of tasks~\cite{wu2023medical,xie2024pasam,zheng2024tuning,GONG2024103324}. The LoRA~\cite{hu2022lora} is a parameter efficient fine-tuning method technique that introduces a trainable linear projection layer within the Transformer architecture. This helps restore SAM's ability to extract high-level context features. Unlike Adapter, which works well for small datasets, LoRA incorporates a low-rank matrix to represent the weight change, reducing both the reliance on the pre-trained model and the risk of overfitting during fine-tuning~\cite{zhongconvolution,10637992,YE2024102826,xiao2024segment,wang2024task}. Together, these methods provide powerful ways to adapt SAM to specialized tasks while maintaining its generalization capabilities.

\section{The Proposed Method}

\subsection{Preliminary}

%First, we briefly recap the design of SAM and LoRA. SAM comprises three key modules: (a) Image encoder: a heavy ViT-based~\cite{dosovitskiy2021an} backbone for image feature extraction. (b) Prompt encoder: encoding the prompt positional information, such as points, boxes, and masks. (c) Mask decoder: a lightweight Transformer-based decoder that concatenates image embedding and prompt tokens for final mask prediction. The released SAM model is trained on a very large-scale dataset, which endows SAM with powerful prior knowledge and zero-shot capabilities. However, it is necessary to fine-tune SAM to be injected with domain-specific guidance information to achieve high-quality segmentation in RSVMO.

We first briefly recap the design of SAM and LoRA. SAM consists of three main modules: (a) Image Encoder: a heavy ViT-based~\cite{dosovitskiy2021an} backbone for image feature extraction, (b) Prompt Encoder: encoding the prompt positional information, such as points, boxes, and masks, and (c) Mask Decoder: a lightweight Transformer-based decoder that concatenates image embedding and prompt tokens for final mask prediction. The SAM model is trained on a vast large-scale dataset, which endows SAM with powerful prior knowledge and zero-shot capabilities. 

LoRA is an efficient method for injecting domain-specific knowledge by freezing the pre-trained model weights and introducing small trainable decomposition matrices into each layer of the Transformer block. Specifically, given a pretrained weight matrix $W_0 \in \mathbb{R}^{m\times n} $ of SAM, LoRA adds a pair of encoder $W_e \in \mathbb{R}^{r\times n}$ and decoder $W_d  \in \mathbb{R}^{m\times r} $, which represent rank decomposition matries, and $ r\ll \min(m,n) $. With LoRA, the original SAM forward propagation changes from $h = W_0x$ to
\begin{equation}
h = W_{0}x+W_dW_{e}x.
  \label{eq:loar1}
\end{equation}

\subsection{ROS-SAM}
An overview of the proposed method is depicted in Figure~\ref{fig:2}. The upper part outlines the training process of ROS-SAM. First, remote sensing images with varying sizes are fed into the image encoder using LSJ and random rotations. In the image encoder, we update the parameters of LoRA and the network's last block to incorporate remote sensing domain-specific guidance and enhance feature discrimination. Next, we exploit the HQ mask decoder, an extension of the original decoder that integrates multi-stage image features, prompt tokens, and mask tokens to produce high-quality mask predictions. Meanwhile, we introduce alternating updates to optimize the original mask decoder and the HQ mask decoder. The lower part describes the inference process. Initially, the image is cropped based on prompt information, ensuring that only one object is centered for inference. We then upsample the patches once to enlarge the features for higher-quality inference. Finally, the individual inference patches are restored to their original positions to align with the initial image, ensuring consistency with the original input.

\subsubsection{Fine-tuning the image encoder}\label{sec:321}

\begin{figure}
  \centering
  \includegraphics[width=0.85\linewidth]{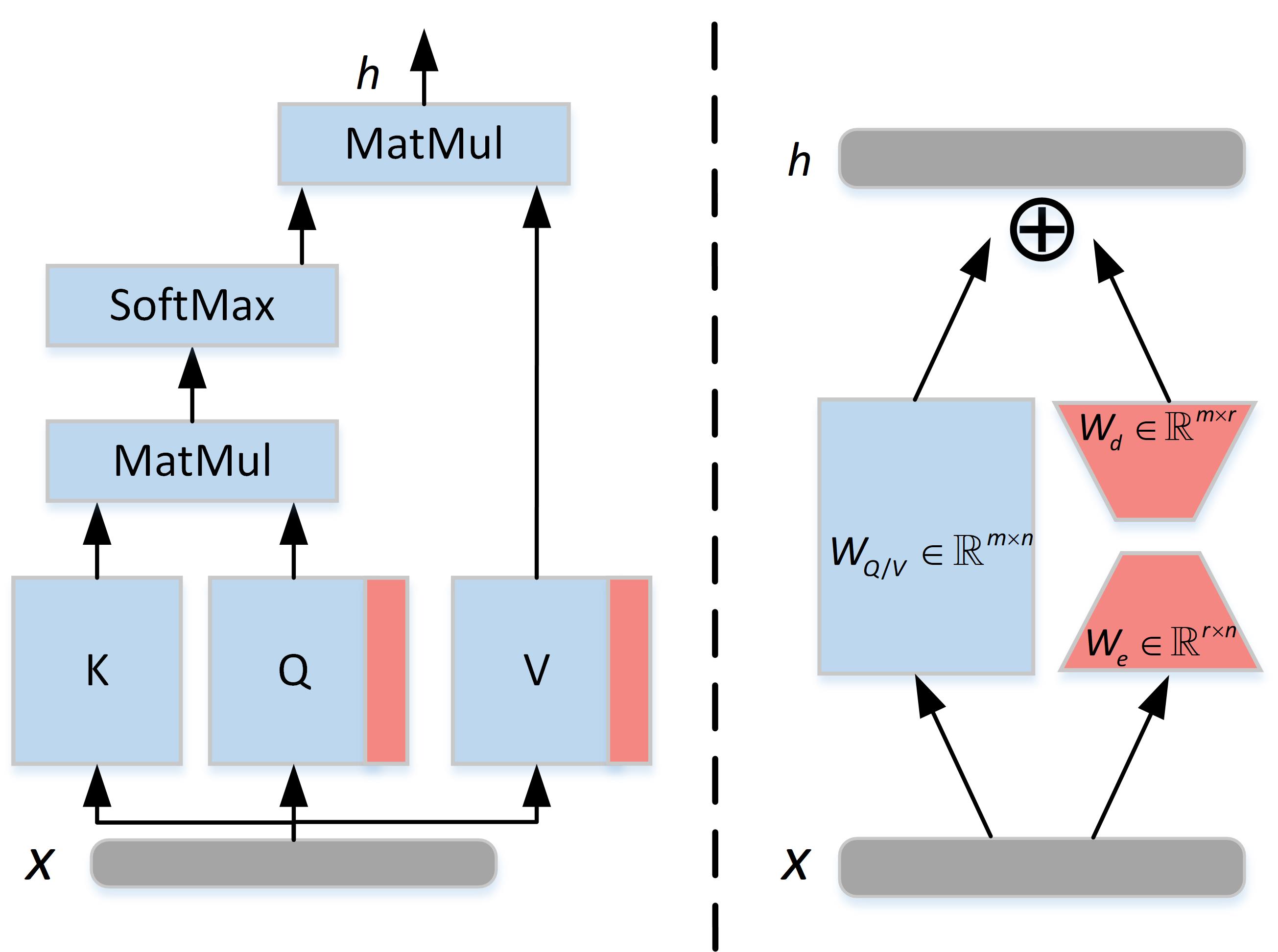}
  \caption{Illustration of introducing LoRA. Specifically, we inject low-rank decomposition matrices into the pre-trained Query and Value projection matrices of each self-attention layer.}
  \label{fig:3}
    \vspace{-15pt}  
\end{figure}

\begin{figure}
  \centering
  \includegraphics[width=0.85\linewidth]{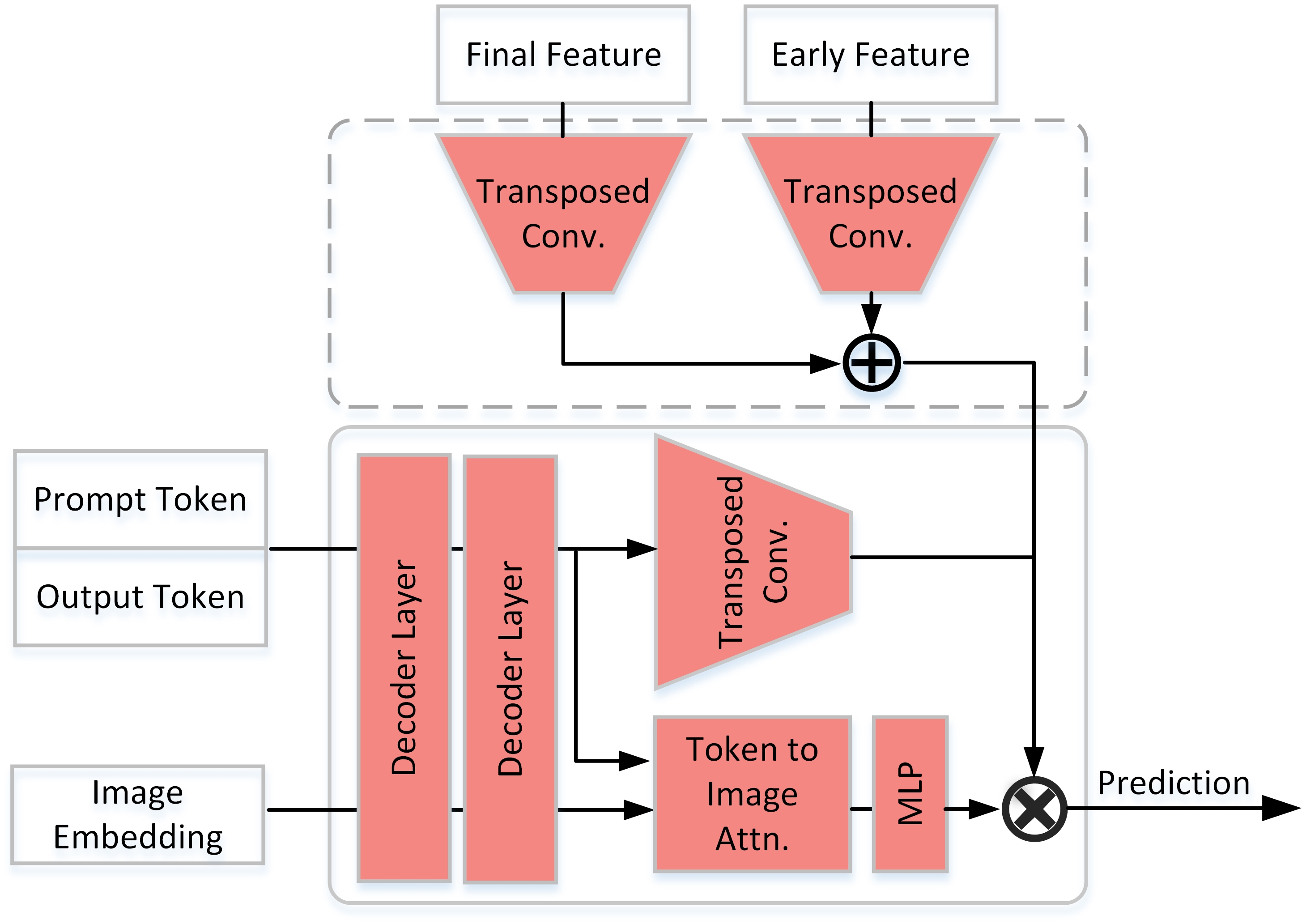}
  \caption{Illustration of the mask decoder. The solid box shows the mask decoder of the original SAM, and then splicing on the dotted box part is the HQ mask decoder.}
  \label{fig:4}
 \vspace{-15pt}  
\end{figure}

The image encoder of SAM is a heavy ViT with powerful prior knowledge. To adapt it for the RSVMO while preserving as much of this prior knowledge as possible, the key challenge lies in effectively injecting domain-specific information. To address this, we introduce LoRA to fine-tune all Transformer layers of the image encoder, as illustrated in Figure~\ref{fig:3}. 

In the image encoder, the input image is first processed through patch embedding and then passed through multiple cascaded Transformer blocks within the ViT. Each Transformer block progressively captures dependencies between image patches using the self-attention mechanism. To inject domain-specific knowledge without compromising the model’s original capabilities, we apply LoRA at the attention calculation stage. Specifically, we introduce a new branch parallel to the Query (Q) and Value (V) matrices. This branch decomposes the original Q and V matrices into low-rank matrices using an encoder-decoder structure, implemented via simple linear layers. During training, only the weights of the low-rank matrices are updated, allowing the model to quickly adapt to new tasks while preserving the generalizable features learned during pre-training. Based on previous experience~\cite{hu2022lora}, we find that updating the Q and V matrices is the most effective way to balance model performance and computational efficiency.

Additionally, we observe that the model often struggles when multiple significant objects are present within the prompt box, such as airplanes and boarding bridges. SAM is a segmentation model without explicit semantic understanding, and it tends to segment out all prominent objects, regardless of their relevance. Research~\cite{Zhu_2021_CVPR,Ji_2022_CVPR} has shown that in semantic segmentation models, shallow layers typically capture texture details, while deeper layers encode richer global context information. Building on this insight, we update the last block of the image encoder to extract more discriminative features during the training, improving the model's ability to focus on the relevant objects in complex scenes.

\subsubsection{Mask decoder and high-quality mask decoder}\label{sec:322}

As shown in the solid box of Figure~\ref{fig:4}, the mask decoder of the original SAM uses two Transformer layers to output the token that is adopted for mask prediction. This token predicts dynamic MLP weights and then performs point-wise products with the mask features. HQ-SAM~\cite{NEURIPS2023_5f828e38} believes that an effective mask decoder should integrate both high-level object context and low-level edge information to achieve high-quality predictions. The different stages of the image encoder capture distinct types of information: the early layers typically extract local fine-grained features, while the later layers focus on global context. Based on this, we use the HQ-SAM mask decoder to generate high-quality masks, as illustrated in Figure~\ref{fig:4}. Another key advantage of using the HQ-SAM mask decoder is that it is trained on the HQSeg-44K dataset, which contains 44,320 highly accurate mask annotations, providing strong prior knowledge of edge details.

We note two important factors: First, SAM's mask decoder is a relatively lightweight network with numerous upsampling operations involving convolutions and transposed convolutions. These operations are designed for pixel-level categorization, and updating this portion of the model does not lead to catastrophic forgetting of the prior knowledge. Second, the inductive bias inherent in convolutions and transposed convolutions requires parameter tuning for new tasks. Therefore, unlike HQ-SAM, we propose updating these weights during training. Specifically, we alternately update the SAM mask decoder and the HQ-SAM mask decoder during training to refine both components simultaneously.

\subsubsection{Training and inference of ROS-SAM}\label{sec:323}

%\textbf{ROS-SAM training.} The image of RSVMO has very significant differences from natural images. RSVMO images are usually larger in size, have more different object scales, and lack direction. In response to the above characteristics, we redesigned the training data pipeline. First, we use large scale jittering~\cite{ghiasi2021simple} to improve the model's generalizability to different object scales. Second, since remote sensing targets are not affected by gravity and do not have up and down orientation, it makes sense for us to use random rotations to add more training samples. During the training process, we fixed the model parameters of the pre-trained SAM while making the red parts in Figure~\ref{fig:2} learnable. We supervise mask prediction tokens with a combination of both BCE Loss and Dice Loss, and we take turns updating the network weights from SAM Mask and ROS-SAM Mask. We trained our ROS-SAM for 24 epochs using a learning rate of 1e-3.

\textbf{ROS-SAM training.} The images of RSVMO exhibit significant differences from natural images. They are typically larger in size, contain objects at varying scales, and lack a clear orientation. To address these characteristics, we redesign the training data pipeline. First, we utilize large-scale jittering~\cite{ghiasi2021simple} to enhance the model's ability to generalize across different object scales. Second, since remote sensing targets are not affected by gravity and lack fixed orientations, we introduce random rotations to generate more diverse training samples. During training, we freeze the parameters of the pre-trained SAM, making only the red components in Figure~\ref{fig:2} learnable. We supervise the mask prediction tokens using a combination of both Binary Cross-Entropy (BCE) Loss and Dice Loss, and alternately update the network weights between the SAM Mask and ROS-SAM Mask. We trained our ROS-SAM for 24 epochs with a learning rate of 1e-3.

\textbf{ROS-SAM inference.} The inference pipelines used by SAM, HQ-SAM, and other works, which rely on direct resizing, are not suitable for remote sensing images, which typically have large sizes and small objects. To achieve refined predictions, we redesign the inference pipeline. First, we crop N×512×512 patches near the object based on the location information of the prompt. Next, we use bicubic interpolation to upsample the patches to N×1024×1024. These patches are then sequentially input into ROS-SAM and stored in memory. Finally, the patches are restored to the full mask based on the positional information of the prompt. Our experiments show that upsampling by a factor of two yields optimal results, while excessive upsampling leads to a significant drop in accuracy.

\section{Experiments}
\subsection{Experimental Setup}
%\textbf{Datasets.} We use the SAT-MTB~\cite{10130311} for training and evaluation, which is the only dataset available for remote sensing video that includes tracking, detection, and segmentation. The SAT-MTB dataset contains 249 videos totaling about 50,000 frames and includes the four most common types of ground targets: airplanes, cars, ships, and trains. In fact, cars don't need to generate masks, they usually only take up 10 or so pixels and have very uniform shapes that are basically the same as the detection boxes. Besides, some of the labels in the original dataset are missing. Therefore, these two parts of the data we will exclude from the dataset. Given that most frames of a video are highly similar, we randomly sample $\frac{1}{4}$ these frames for each video to generate the final dataset. For more detail as shown in Table~\ref{tab:1}, all subsequent experiments were performed on this dataset if not otherwise noted.

\textbf{Datasets.} We use the SAT-MTB~\cite{10130311} for training and evaluation, which is the only available dataset for remote sensing video that includes object tracking, detection, and segmentation tasks. The SAT-MTB dataset comprises 249 videos, totaling approximately 50,000 frames, and covers the four most common types of ground targets: airplanes, cars, ships, and trains. Notably, for cars, mask generation is unnecessary, as they typically occupy only about 10 pixels and have uniform shapes that closely match the detection boxes. Additionally, some labels in the original dataset are missing, so we exclude these incomplete data points from our analysis. To mitigate the high frame similarity in most videos, we randomly sample 1/4 of the frames from each video to create the final dataset. Detailed description of the dataset is provided in the supplemental material. Unless otherwise specified, all subsequent experiments are conducted using this modified dataset.

% \begin{table}
%   \centering
%   \begin{tabular}{lc}
%     \toprule
%     Attribute & Number \\
%     \midrule
%     Number of training sets & 8333 \\
%     Number of test sets & 2284 \\
%     Maximum image size & 2160×1080\\
%     Minimum image size & 512×512\\
%     Number of objects & 228367\\
%     \bottomrule
%   \end{tabular}
%   \caption{Details of our used dataset.}
%   \label{tab:1}
% \end{table}

\textbf{Evaluation metrics.} To accurately quantify the performance of the model, we utilize BIoU~\cite{Cheng_2021_CVPR} and IoU to jointly evaluate the prediction accuracy.

\subsection{Ablation Experiments}

\begin{figure}
  \centering
		\includegraphics[width=0.85\linewidth]{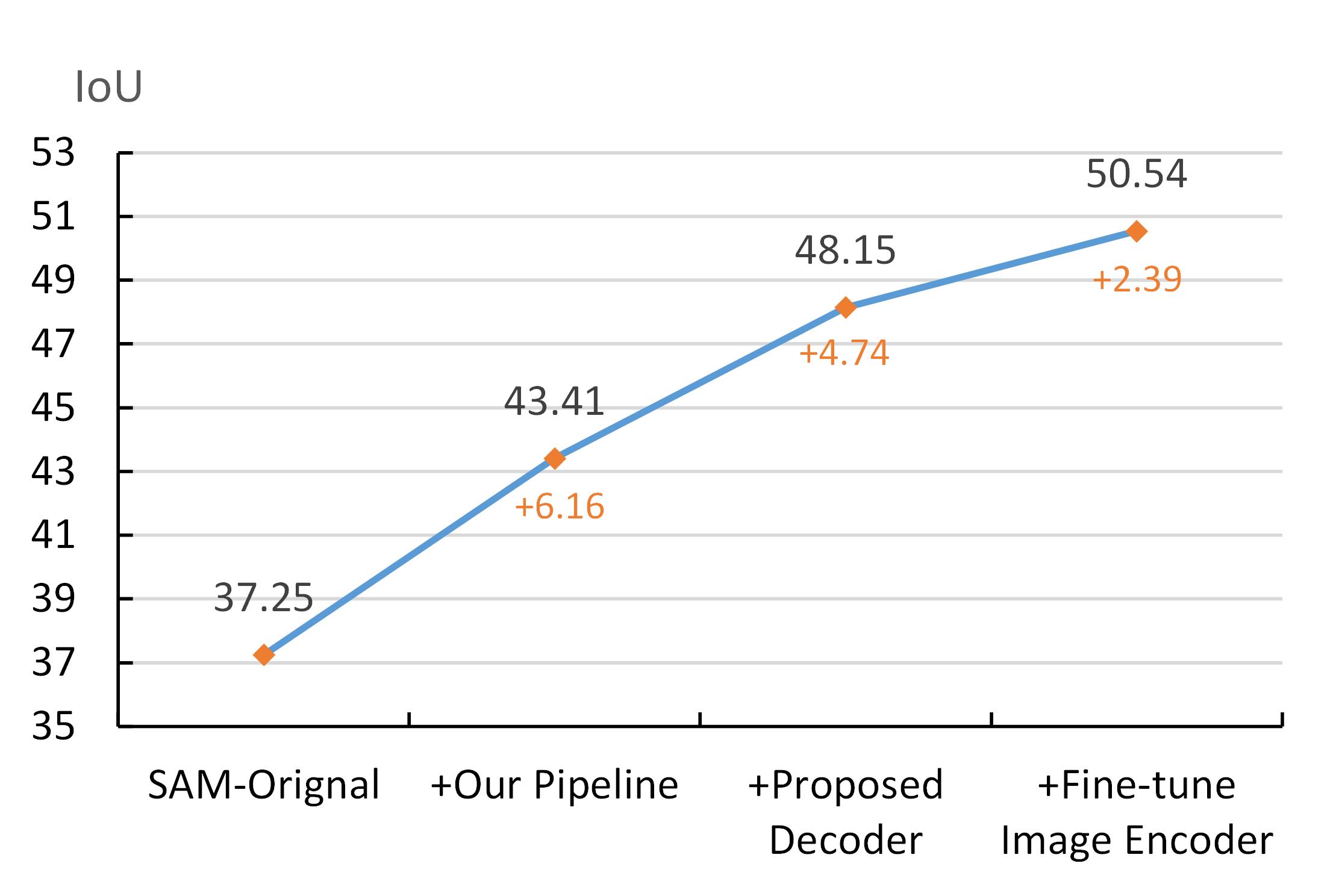}
   \caption{Ablation study of the proposed method. + indicates a new method adds to the previous one, and the orange numbers indicate the amount of growth.}\label{fig:5}
    \vspace{-5pt}  
\end{figure}

We conduct comprehensive ablation studies of each proposed method, using SAM1-L as the baseline, i.e., the image encoder adopts the ViT-L version. All experimental settings remain consistent, except for the validation approach. As shown in Figure~\ref{fig:5}, each of our methods leads to a significant improvement in prediction accuracy.

\begin{figure*}
  \centering
		\includegraphics[width=0.9\linewidth]{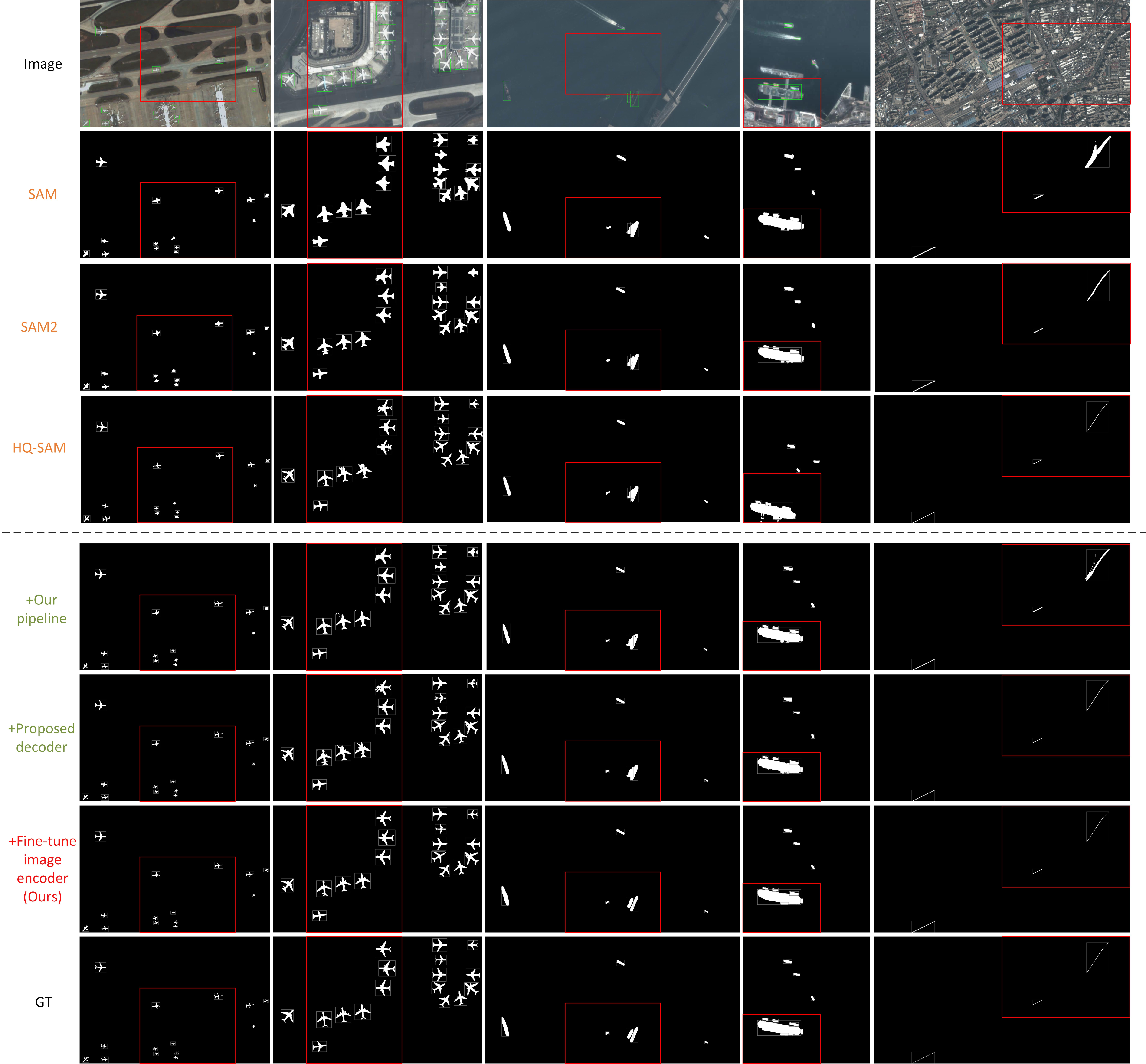}
   \caption{Visualized results of our proposed ROS-SAM and compared SOTA methods. We show the results of the ablation experiment (bottom) and the comparison experiment (top) in one image. + indicates a new method adds to the previous one.}\label{fig:ab}
 \vspace{-15pt}  
   
\end{figure*}

In Figure~\ref{fig:ab}, we visualize the prediction results, highlighting how our methods enhance the model's performance. For instance, our pipeline enables the model to focus on individual objects, resulting in extraction outcomes that are much closer to the true shape of the objects, such as the predicted airplane in the second column. Moreover, when the proposed decoder is incorporated, the model’s predictions become more refined. The decoder effectively learns part of the RSVMO knowledge, improving the model's ability to discriminate between objects. This is evident in the fifth column, where SAM fails to predict the train, and the predicted shape is much larger than the train itself, likely misidentifying the railroad instead. This discrepancy can be attributed to SAM’s limited domain knowledge of RSVMO. In contrast, our ROS-SAM achieves the best results, with predictions that are nearly identical to the ground truth. Notably, ROS-SAM successfully differentiates between an airplane and a boarding bridge, which all other models fail to do. Additionally, ROS-SAM delivers highly accurate predictions that closely match the true shape of objects, even when distinguishing between objects in close proximity, as shown in the third and fourth columns. While ROS-SAM excels overall, it does have some limitations. For example, it struggles to accurately differentiate the number of airplane engines and may not predict the shape of objects that are difficult to distinguish visually (where manual annotations based on prior knowledge may be more accurate). For further visual comparisons, please refer to the supplemental materials.

 \begin{table}
  \centering
  \begin{tabular}{lcc}
    \toprule
    Method & IoU &BIoU \\
    \midrule
    SAM~\cite{Kirillov_2023_ICCV} & 37.25 & 37.14\\
    +inference pipeline & \textbf{43.41} & \textbf{43.30} \\
    \midrule    
    SAM2~\cite{ravi2024sam} & 36.80  &  36.67\\
    +inference pipeline & \textbf{41.75} & \textbf{41.55} \\
    \bottomrule
  \end{tabular}
  \caption{Ablation study of our inference pipeline.}
  \label{tab:2}
    \vspace{-15pt}  
\end{table}

\subsubsection{Ablation study of our pipeline}
%Our pipeline includes the training pipeline and inference pipeline. First, we verify the validity of the inference pipeline in two models, SAM and SAM2, which are not trained. As shown in Table~\ref{tab:2}, only using our inference pipeline can improve IoU by over 5\%. Second, we also test the effect of different sampling algorithms, sampling rates, and whether only one object is inferred during the inference stage. As shown in Table~\ref{tab:3}, we use the version without sampling, i.e., cropped to 1024×1024 and containing multiple objects as the baseline. The experiment results show that the above three aspects have some degree of effect on the accuracy of the algorithm. Setting the sampling rate to 2 is the most appropriate, with too high upsampling resulting in the edges of the object behaving in a jagged fashion that does not match reality. The bicubic interpolation and only one object during inference likewise all have a positive effect on the algorithm. Subsequently, we all use the current best configuration as an inference pipeline.

First, we validate the effectiveness of the inference pipeline using two models, SAM and SAM2, which have not been trained. As shown in Table~\ref{tab:2}, applying our inference pipeline for SAM results in an improvement of over 6\% in both IoU and BIoU. For SAM2, our inference pipeline can also bring significant improvements. Next, we evaluate the impact of various factors during inference, including different sampling algorithms, sampling rates, and the effect of inferring only one object at a time. The results are shown in Table~\ref{tab:3}, the baseline configuration involves no sampling (cropped to 1024×1024 and containing multiple objects ). The experiments demonstrate that all three factors have a noticeable impact on the accuracy. Among the different sampling rates, a rate of 2 proves to be the most effective, as higher upsampling rates result in jagged object edges that do not reflect the true shape of the objects. Additionally, both bicubic interpolation and limiting inference to a single object also improve the performance. Based on these findings, we adopt the best-performing configuration as the final inference pipeline.

 \begin{table}
  \centering
  \begin{tabular}{lcc}
    \toprule
    Method & IoU &BIoU \\
    \midrule
    Baseline &39.16 & 39.04 \\
    Sampling rate = 2, bilinear  &41.58 & 41.45\\
    Sampling rate = 2, bicubic  &42.17 & 42.13\\
    \midrule
    Sampling rate = 4, bilinear  &40.37 & 40.11\\    
    Sampling rate = 4, bicubic  &40.81 & 40.43\\
    \midrule
    Sampling rate = 2, bicubic, only one &\textbf{43.41} &\textbf{43.30}\\
    Sampling rate = 4, bicubic, only one & 41.73&41.51\\     
    \bottomrule
  \end{tabular}
  \caption{Ablation study of the effects of different settings on inference pipeline.}
  \label{tab:3}
   \vspace{-5pt}  
\end{table}

%The training pipeline is equally crucial to the algorithm's results. To validate the effectiveness of the proposed training pipeline, we conduct ablation experiments in the proposed ROS-SAM. Our training pipeline introduces two data augmentations LSJ (from 0.1 to 4.0) and random rotation. As shown in Table~\ref{tab:4}, both random rotation and LSJ improve model accuracy, with LSJ improving more significantly due to the introduction of more scaled objects, and random rotation essentially just increasing the training samples and therefore improving is limited.

We then conduct ablation experiments for the training pipeline of ROS-SAM. The proposed training pipeline incorporates two data augmentation techniques: LSJ (which scales objects from 0.1 to 4.0) and random rotation. As shown in Table~\ref{tab:4}, both augmentations contribute to improved model accuracy, with LSJ providing a more significant boost. This is because LSJ introduces a wider range of object scales, allowing the model to better generalize. In contrast, random rotation primarily increases the number of training samples, resulting in a more limited improvement. 

 \begin{table}
  \centering
  \begin{tabular}{lcc}
    \toprule
    Method & IoU &BIoU \\
    \midrule
    ROS-SAM                 &49.19 & 49.05 \\
    +random rotation         &49.43 & 49.26\\
    +LSJ                     &50.03 & 49.89\\
    +LSJ and random rotation &\textbf{50.54} & \textbf{50.36}\\
    \bottomrule
  \end{tabular}
  \caption{Ablation study of our training pipeline.}
  \label{tab:4}
   \vspace{-5pt}  
\end{table}

 \begin{table}
  \centering
  \begin{tabular}{lcc}
    \toprule
    Method & IoU &BIoU \\
    \midrule
    SAM                        &43.41 & 43.30 \\
    Update Mask Decoder        &42.82 &42.69\\
    Update HQ Mask Decoder      &47.15&47.11\\
    Alternately Update Two Decoders &\textbf{48.16} & \textbf{48.03}\\
    \bottomrule
  \end{tabular}
  \caption{Ablation study of our proposed mask decoder.}
  \label{tab:5}
      \vspace{-15pt}  
\end{table}

\subsubsection{Ablation study of the mask decoder}
%We improve the mask decoder to generate high-quality prediction results, which includes two main improvements: (1) using the HQ mask decoder for fusing the image embedding with early texture features. (2) alternately updating the HQ mask decoder and the original mask decoder to achieve more accurate pixel prediction. As shown in Table~\ref{tab:5}, when directly updating the mask decoder of SAM, it will lead to a decrease in accuracy. The reason for this is that the original decoder has already been trained on the massive dataset, and directly updating the weights will lead to the appearance of destroying the original prior knowledge. In contrast, the HQ mask decoder is a newly introduced lightweight component and incorporates more stages of image features, so its improvement in accuracy is significant. In contrast, the HQ mask decoder is a newly introduced lightweight component that incorporates multi-stage features, so it is a significant improvement in prediction accuracy. As described in Section~\ref{sec:322}, updating these two components alternately is an optimal option, and they benefit mutually from each other's knowledge.

The proposed mask decoder includes two contributions: 1) using the HQ mask decoder to fuse image embeddings with early texture features, and 2) alternately updating the HQ mask decoder and the original mask decoder to achieve more accurate pixel-level predictions. As shown in Table~\ref{tab:5}, directly updating the mask decoder of SAM leads to a decrease in accuracy. This is because the original decoder has already been trained on a large dataset, and directly adjusting its weights can disrupt the valuable prior knowledge it has learned. In contrast, the HQ mask decoder is a newly introduced lightweight component that integrates multiple stages of image features, resulting in a significant improvement. As discussed in Section~\ref{sec:322}, alternately updating both the HQ mask decoder and the original decoder proves to be the optimal approach, as each benefits from the other's knowledge, leading to better overall performance.

\subsubsection{Ablation study of the image encoder}

%The image encoder is critical to the performance of the model, and we fine-tuned the image encoder to adapt to the object in RSVMO and extract discriminative features. In this ablation experiment, we focus on verifying the performance improvement resulting from fine-tuning the image encoder based on LoRA and fine-tuning the last layer of the image encoder. The experimental results are shown in Figure~\ref{fig:6}, we verified the importance of fine-tuning the image encoder in both SAM and ROS-SAM . It can be seen that LoRA's boost to both models is very large, and since SAM has not undergone any fine-tuning, its improvement will be even more pronounced.

\begin{figure}
  \centering
		\includegraphics[width=1.0\linewidth]{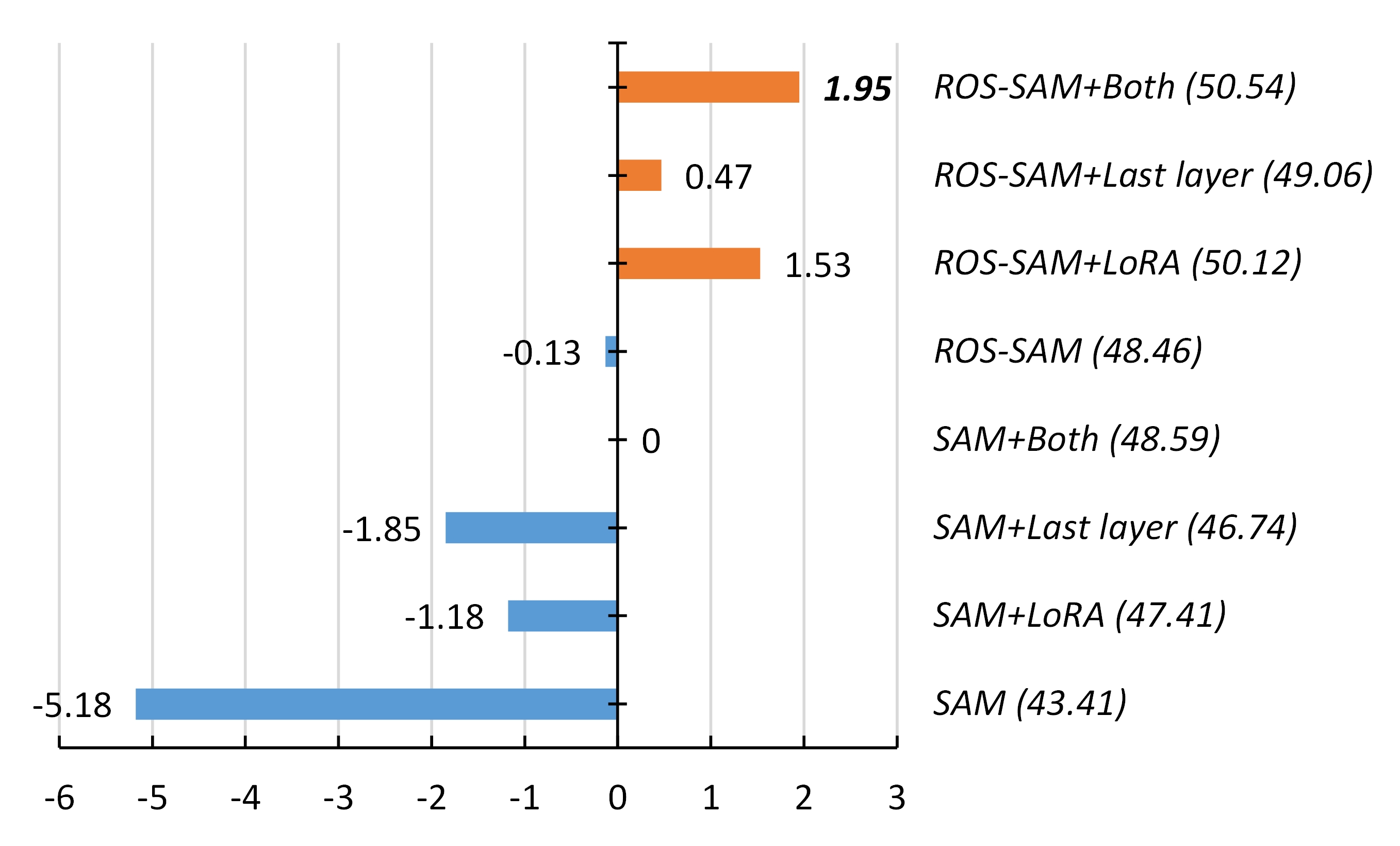}
   \caption{Ablation study of the different fine-tune methods, where SAM+Both is the baseline.}\label{fig:6}

\end{figure}

In this ablation experiment, we specifically evaluate the performance gains achieved by fine-tuning the image encoder using LoRA, as well as by fine-tuning only the last layer of the encoder. The experimental results, shown in Figure~\ref{fig:6}, highlight the significant performance improvements from fine-tuning the image encoder in both SAM and ROS-SAM.

\subsection{Comparison with Other Methods}

We conduct comparative experiments to fully validate the superiority of the ROS-SAM. As shown in Table~\ref{tab:6}, We perform a careful comparison with three algorithms, SAM, SAM2, and HQ-SAM, for which we use our inference pipeline for training and inference. We also fine-tune some other SOTA algorithms, such as~\cite{Chen_2023_ICCV,kim2024cad,xiong2024sam2,hu2023efficiently}. However, they all produce very poor results (the accuracy of the IoU is less than the original SAM), so we do not continue with the comparisons. Figure~\ref{fig:ab} only shows the most representative visualizations, we analyze the results of a large number of visualizations and summarize them as follows: (1) Models like SAM1 and SAM2 that have not been fine-tuned are not capable of high-quality inference, they have very fuzzy edges most of the time. (2) Although HQ-SAM generates refined prediction results, it is not semantically discriminative, so there are more misclassifications. (3) When the features are clear, the results predicted by our method are almost perfect, and when the features are fuzzy, the annotator can predict the approximate shape of the target, which is difficult to achieve with our method.

 \begin{table}
  \centering
  \begin{tabular}{lcc}
    \toprule
    Method & IoU &BIoU \\
    \midrule
    SAM~\cite{Kirillov_2023_ICCV}   &43.41 (37.25) & 43.30 (37.14)\\
    SAM2~\cite{ravi2024sam}        &41.75  (36.80) &41.55  (36.67)\\
    HQ-SAM~\cite{NEURIPS2023_5f828e38}  &47.15 (43.27)&47.11 (43.21)\\
    Ours &\textbf{50.54}&\textbf{50.36}\\
    \bottomrule
  \end{tabular}
  \caption{Comparison experiments with other SOTA algorithms. Results of inference according to the original algorithm configuration are shown in parentheses.}
  \label{tab:6}
   \vspace{-15pt}  
\end{table}

% \begin{figure*}
%   \centering
% 		\includegraphics[width=1.0\linewidth]{Compare}
%    \caption{Visualized results of different methods. To give a complete display of the predictive quality of our method, the images here are uncropped. }\label{fig:comp}
% \end{figure*}

\subsection{Experiment Results on Other Datasets }

\subsubsection{Segmentation results on static image dataset}

\begin{table}[bht]
 \centering
\begin{tabular}{l|cc|cc}
    \toprule
    & \multicolumn{2}{c|}{iSAID} & \multicolumn{2}{c}{NPWS VHR-10} \\
    \cmidrule(r){2-3} \cmidrule(l){4-5}
    \multirow{-2}{*}{Method} & \multicolumn{1}{c}{IoU} & \multicolumn{1}{c|}{BIoU} & \multicolumn{1}{c}{IoU} & \multicolumn{1}{c}{BIoU} \\
    \midrule
    SAM      & 53.19  & 48.86  & 65.54 & 68.44 \\
    SAM2     & 55.79  & 52.34  & 68.84   & 69.18   \\
    HQ-SAM   & 63.96  & 57.48  & 78.44 & 76.10 \\
    Ours     & \textbf{73.22}  & \textbf{68.22}  & \textbf{87.46} & \textbf{84.56} \\
    \bottomrule
\end{tabular}
  \caption{The results on static remote sensing datasets.}\label{tab:r}
\vspace{-10pt}
\end{table}

We have conducted experiments on two static datasets to further verify the effectiveness and generalization of the proposed method. iSAID~\cite{waqas2019isaid} and NPWS VHR-10~\cite{rs12060989} datasets are representative instance segmentation datasets, comprising 15 and 10 common types of objects, respectively. As shown in Table~\ref{tab:r}, our method is also applicable to static images and has more advantages. 

%Zero-shot capabilities of the model are critical to our task, and we tested the model's performance on multiple remote sensing object tracking datasets, including SatSOT~\cite{9672083}, VISO~\cite{9625976}, OOTB~\cite{9625976}. These datasets differ from the training sets in terms of resolution, sensor, time, and place of capture, which can effectively validate the generalization ability of the model. Considering that the tracking dataset does not have segmentation masks, we only performed a qualitative evaluation and no longer quantitatively analyzed the predictions, as shown in Figure~\ref{fig:zeroshot}. It can be seen that RSVMO still shows superior performance on these datasets, generating high-quality labels.

\subsubsection{Zero-shot segmentation on object tracking}

\begin{figure}
  \centering
		\includegraphics[width=1.0\linewidth]{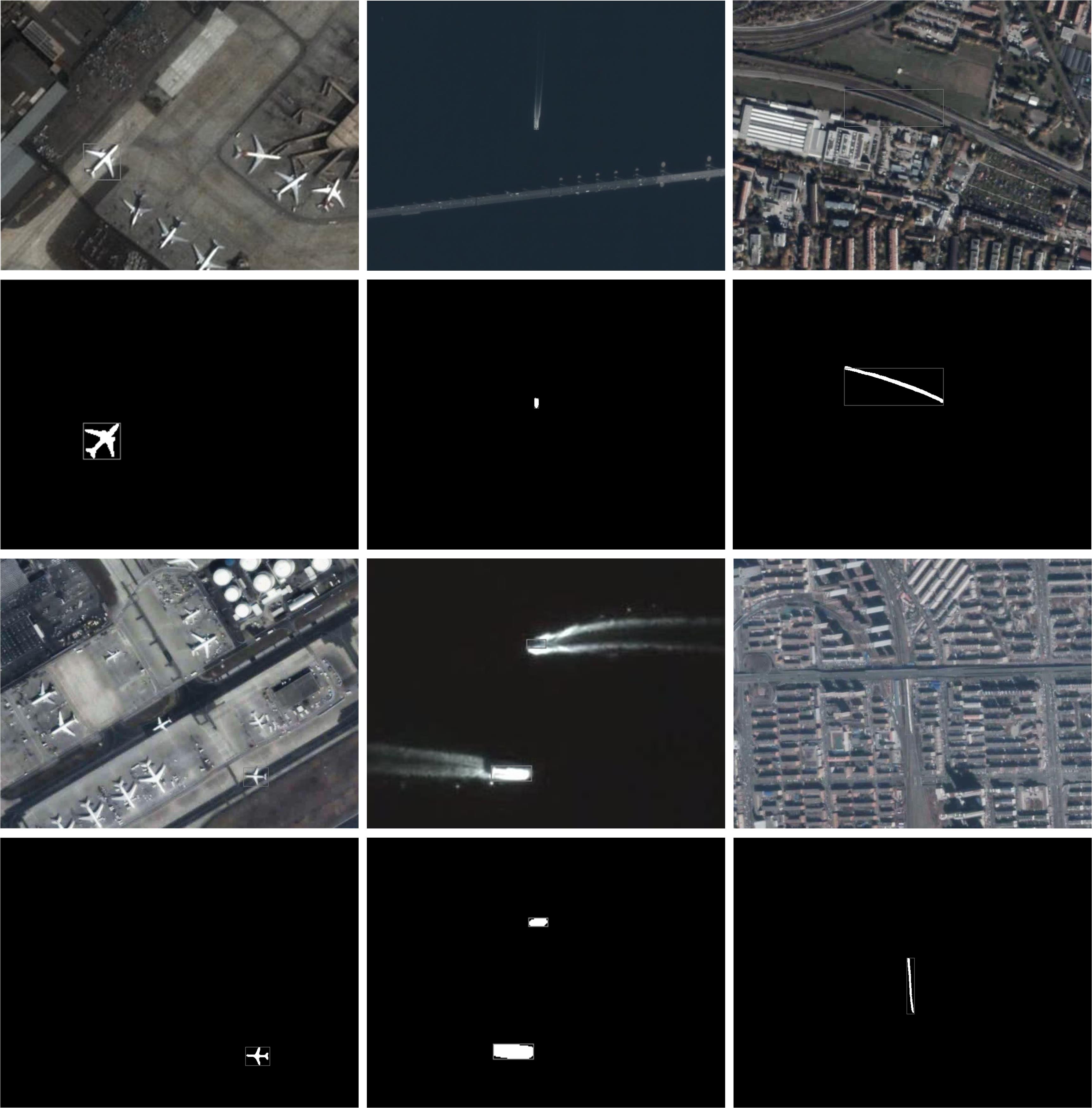}
   \caption{Results on remote sensing object tracking dataset.}\label{fig:zeroshot}
    \vspace{-15pt}  
\end{figure}
The zero-shot capabilities of the model are crucial for our task, and we evaluate its performance on several remote sensing object tracking datasets, including SatSOT~\cite{9672083}, VISO~\cite{9625976}, and OOTB~\cite{9625976}. These datasets differ from the training set in terms of resolution, sensor type, and the time and location of capture, providing a robust test of the model's generalization ability. Since the tracking datasets lack segmentation masks, we conduct a qualitative evaluation rather than a quantitative analysis of the predictions. As shown in Figure~\ref{fig:zeroshot}, ROS-SAM demonstrates superior performance across these datasets, consistently generating high-quality segmentation masks. Due to the length limitation of the paper, more visualization results refer to the supplemental materials.

\section{Conclusion}

%In this paper, we present ROS-SAM, a model built upon SAM with improvements to three key components: the data pipeline, image encoder, and mask decoder, designed to achieve high-quality object prediction in RSVMO. Ablation and comparison experiments demonstrate that these components significantly enhance SAM's performance in RSVMO scenarios. Furthermore, the generalization capability of ROS-SAM is validated through cross-data experiments. We envision that ROS-SAM will serve as a powerful tool for advancing fine-grained segmentation in remote sensing video analysis.

In this paper, we present ROS-SAM, a model built upon SAM with contributions to three critical components: the data pipeline, image encoder, and mask decoder. These components are specifically designed to achieve high-quality object mask prediction in remote sensing moving object segmentation. Ablation studies and comparative experiments show that these modifications substantially enhance SAM's performance in RSVMO tasks. Additionally, the generalization ability of ROS-SAM is validated through cross-dataset experiments. We anticipate that ROS-SAM will become a powerful tool for advancing fine-grained segmentation in remote sensing video analysis.

\noindent\textbf{Acknowledgement.} This work was supported by the National Natural Science Foundation of China (No. 62402354, No. 62362023), and the Key Research and Development Project of Hainan Province (ZDYF2024GXJS262), and Hainan Province Science and Technology Special Fund (ZDYF2024GXJS313).

{
    \small
    \bibliographystyle{ieeenat_fullname}
    \bibliography{main}
}
\clearpage
\setcounter{page}{1}
\maketitlesupplementary

\section{The details of our used dataset}
\label{sec:rationale}
% 

% Having the supplementary compiled together with the main paper means that:
% % 
% \begin{itemize}
% \item The supplementary can back-reference sections of the main paper, for example, we can refer to \cref{sec:intro};
% \item The main paper can forward reference sub-sections within the supplementary explicitly (e.g. referring to a particular experiment); 
% \item When submitted to arXiv, the supplementary will already included at the end of the paper.
% \end{itemize}
% % 
% To split the supplementary pages from the main paper, you can use \href{https://support.apple.com/en-ca/guide/preview/prvw11793/mac#:~:text=Delete%20a%20page%20from%20a,or%20choose%20Edit%20%3E%20Delete).}{Preview (on macOS)}, \href{https://www.adobe.com/acrobat/how-to/delete-pages-from-pdf.html#:~:text=Choose%20%E2%80%9CTools%E2%80%9D%20%3E%20%E2%80%9COrganize,or%20pages%20from%20the%20file.}{Adobe Acrobat} (on all OSs), as well as \href{https://superuser.com/questions/517986/is-it-possible-to-delete-some-pages-of-a-pdf-document}{command line tools}.

\begin{figure}[h!]
  \centering
		\includegraphics[width=1.0\linewidth]{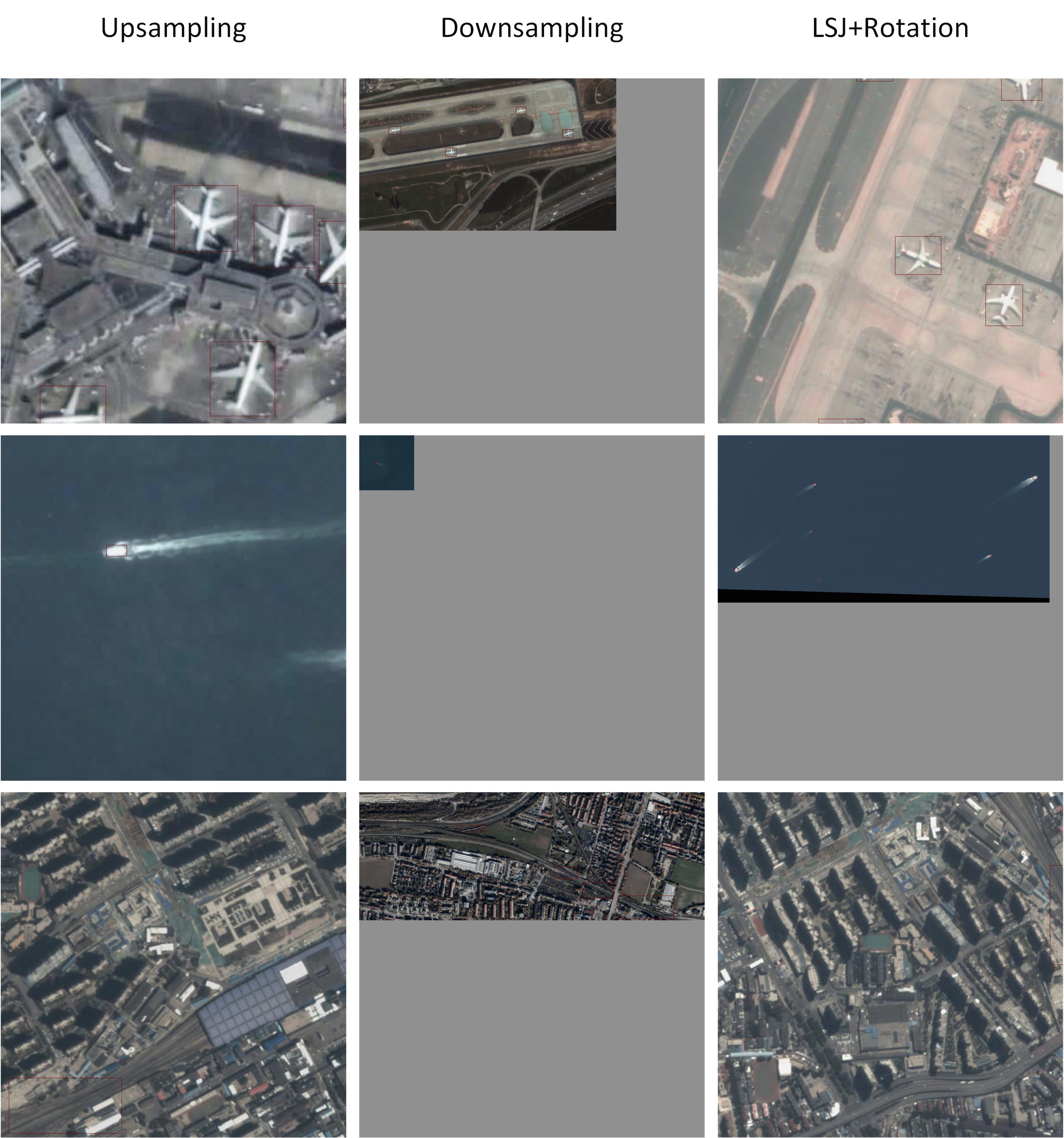}
   \caption{Visualization of partial training samples.}\label{fig:sup1}
   \vspace{-5pt}
\end{figure}

Details of the dataset used in this study are presented in Table~\ref{tab:sp1}. This dataset poses significant challenges due to its diverse range of image and object sizes, with the average object size being relatively small (MS COCO defines small objects as those measuring smaller than 32×32 pixels, and the majority of objects in this dataset fall below that threshold.). Additionally, Figure~\ref{fig:sup1} illustrates sample images from the training set, which are generated using large-scale jittering (LSJ) and random rotations. For original data, we resize images randomly within a scale range of 0.1 to 4.0. If the resized dimensions exceed 1024×1024 pixels, the images are cropped to fit; if they are smaller, gray borders are added to the bottom-right corner to pad the image. This data augmentation strategy effectively enhances both the diversity and quantity of training samples.

\begin{table}[h!]
  \centering
  \begin{tabular}{lc}
    \toprule
    Attribute & Number \\
    \midrule
    Number of training sets & 8,333 \\
    Number of test sets     & 2,284 \\
    Maximum image size      & 2160×1080\\
    Minimum image size      & 512×512\\
    Number of objects       & 52,123\\
    Maximum object size     & 8422  \\
    Minimum object size     & 12\\
    Average object size     & 518\\
    Maximum ratio of the object to image & 0.023339 \\
    Minimum ratio of the object to image & 3.111e-6 \\ 
    \bottomrule
  \end{tabular}
  \caption{Details of the dataset used.}
  \label{tab:sp1}
\end{table}

\section{Visualization of experimental results}
\begin{figure*}
  \centering
		\includegraphics[width=1.0\linewidth]{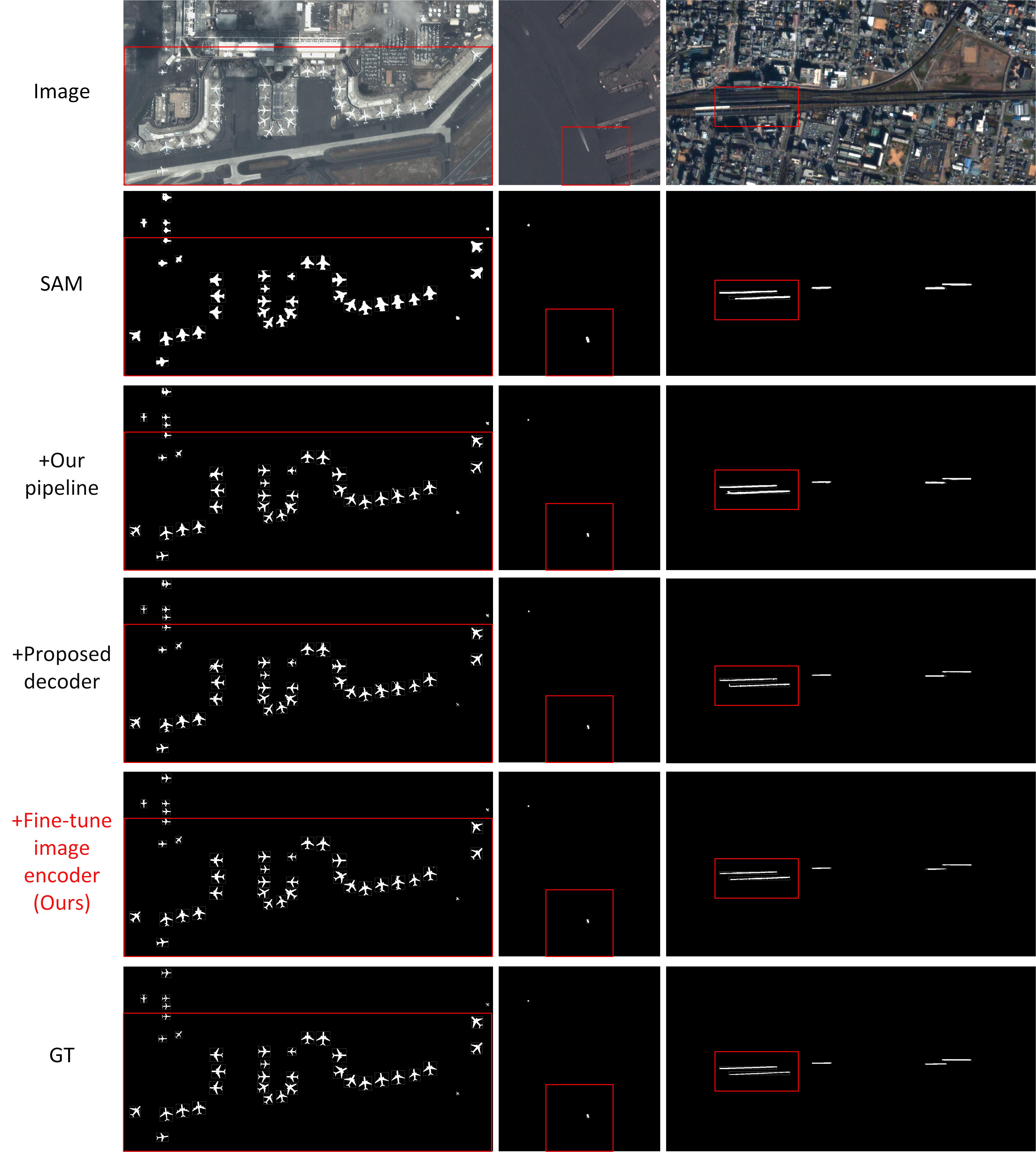}
   \caption{Visualized results of our proposed ROS-SAM. In the second row, the predictions extend beyond the object's actual shape, particularly for the train, where the prediction nearly fills the entire prompt box. In the third row, the predictions lack fine-grained detail and, similar to the second row, fail to distinguish between the foreground and background of the train. In the fourth row, feature discrimination is notably poor, resulting in multiple incorrect predictions; for instance, the boarding bridge and other objects are misclassified as airplanes.}\label{fig:sup2}
\end{figure*}

In Figure~\ref{fig:sup2}, we visualize the prediction results, highlighting how our methods enhance the model's performance. First, a comparison between the results of the original SAM and our pipeline reveals that SAM's predictions are significantly less consistent with the object's actual shape. This inconsistency arises primarily from bias introduced by multiple resizing operations, particularly evident in the ship prediction, where the result extends beyond the prompt box. Second, the proposed decoder improves the object's shape refinement by leveraging early texture information. However, it introduces some degree of feature misjudgment due to limitations in feature interpretation. Finally, when the proposed fine-tuning method is integrated with the image encoder, the resulting predictions are highly accurate and closely align with the ground truth labels.
\section{Zero-Shot segmentation on object tracking}

\begin{figure*}
  \centering
		\includegraphics[width=1.0\linewidth]{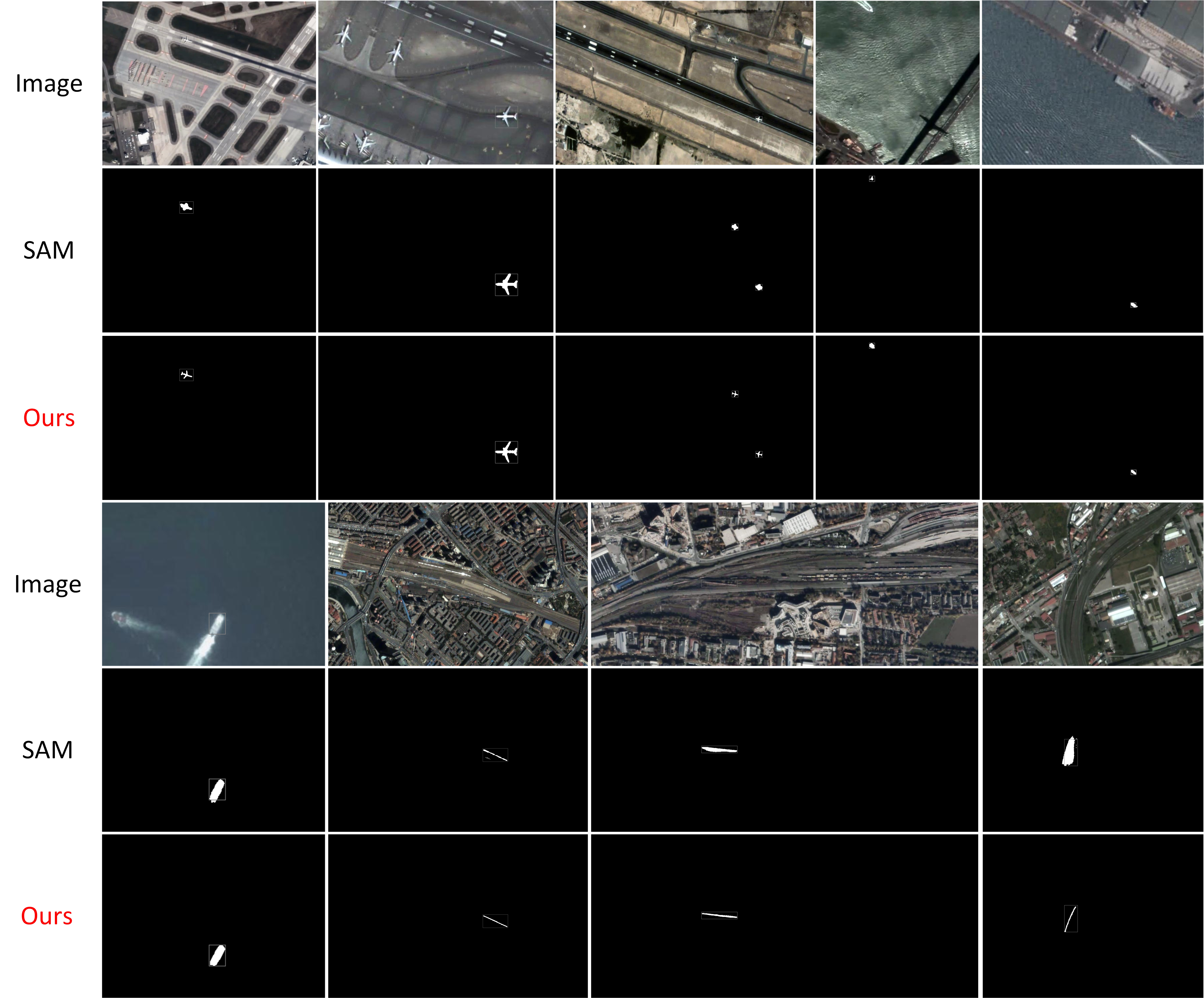}
   \caption{Visualization results of ROS-SAM on the remote sensing object tracking dataset.}\label{fig:ood1}
\end{figure*}

 To evaluate its generalization ability, we test our model on several remote sensing object tracking datasets, as shown in Figures ~\ref{fig:ood1}. Notably, the predictions of our model in this zero-shot setting remain of high quality, demonstrating significantly superior zero-shot capabilities compared to SAM in remote sensing. For airplanes, both our model and SAM can identify airplanes. However, SAM's predictions are highly imprecise, often resembling a four-pointed star rather than the actual shape of an airplane. In contrast, our model achieves fine-grained segmentation, accurately identifying details such as the engine position.  For trains, SAM fails to produce meaningful predictions, whereas our approach shows a clear advantage. For ships, predictions are generally accurate, as the distinct features of ships contrast significantly with the water's surface. Our results highlight challenging scenarios where ships and waves overlap, and our model delivers significantly more accurate predictions in these cases.

% \begin{figure*}
%   \centering
% 		\includegraphics[width=0.8\linewidth]{OOD3}
%    \caption{Visualization results of trains on the remote sensing object tracking dataset.}\label{fig:ood2}
% \end{figure*}

% \begin{figure*}[!t]
%   \centering
% \includegraphics[width=0.8\linewidth]{OOD2}
%    \caption{Visualization results of ships on the remote sensing object tracking dataset.}\label{fig:ood3}
% \end{figure*}


\end{document}